\documentclass[11pt]{article}

\usepackage[preprint]{acl}

\usepackage{times}
\usepackage{latexsym}

\usepackage[T1]{fontenc}

\usepackage[utf8]{inputenc}

\usepackage{microtype}
\usepackage{graphicx}
\usepackage{subcaption}
\usepackage{inconsolata}

\usepackage{graphicx}

\title{Robust Persona-Aware Toxicity Detection with Prompt Optimization and Learned Ensembling}

\author{Berk Atil 
  \\
  Pennsylvania State University \\
  \texttt{bka5352@psu.edu}
  \And
  Rebecca J. Passonneau 
  \\
  Pennsylvania State University \\
  \texttt{rjp49@psu.edu}
   \And
  Ninareh Mehrabi
  \\
  Resolution}

\begin{document}
\maketitle
\begin{abstract}
Toxicity detection is inherently subjective, shaped by the diverse perspectives and social priors of different demographic groups.  While ``pluralistic'' modeling as used in economics and the social sciences aims to capture perspective differences across contexts, current Large Language Model (LLM) prompting techniques have different results across different personas and base models. In this work, we conduct a systematic evaluation of persona-aware toxicity detection, showing that no single prompting method, including our proposed automated prompt optimization strategy, uniformly dominates across all model-persona pairs. To exploit complementary errors, we explore ensembling four prompting variants and propose a lightweight meta-ensemble: an SVM over the 4-bit vector of prompt predictions. Our results demonstrate that the proposed SVM ensemble consistently outperforms individual prompting methods and traditional majority-voting techniques, achieving the strongest overall performance across diverse personas. This work provides one of the first systematic comparisons of persona-conditioned prompting for toxicity detection and offers a robust method for pluralistic evaluation in subjective NLP tasks.
\end{abstract}

\section{Introduction}
The detection of toxic and offensive language is fundamentally a subjective task, deeply influenced by raters' perspectives, social priors, and the social groups targeted in the text \cite{sap-etal-2020-social,sorensen-etal-2025-value}. Further, previous work suggests that each text should be annotated by the group of people who are targeted in the text given demographic differences in toxicity perception \cite{mostafazadeh-davani-etal-2024-d3code,fleisig-etal-2023-majority}. This motivates \textbf{``pluralistic''} toxicity detection that explicitly models toxicity from the perspective of demographic personas: instead of predicting a single majority label, models should be able to approximate how different personas would judge the same content \cite{sorensen-etal-2025-value}. Recent work has explored inference-time prompting to incorporate perspective-taking and other human-centric cues into LLM behavior \cite{xu-etal-2024-walking,dash2025persona}, and LLM-based judges are increasingly used as scalable evaluators for subjective NLP tasks \cite{zheng-etal-2023-judge,liu-etal-2023-g}. However, our systematic evaluation across multiple models shows that a central practical issue remains: \textbf{no single method} consistently dominates across all personas and all underlying base models.

In this work, we study two persona-conditioned prompting approaches and a non-persona prompting method for toxicity detection and show that their strengths are complementary across model--persona pairs. Then, we propose a new persona prompting strategy learned via automatic prompt optimization, using TextGrad-style textual ``differentiation'' to iteratively refine prompts \cite{yuksekgonul-etal-2024-textgrad}. Empirically, this optimized prompting performs comparably to value-profile conditioning \cite{sorensen-etal-2025-value}, but the two methods disagree on a non-trivial subset of examples, suggesting that neither is superior. 

Motivated by the broader success of ensembling and aggregation for improving reliability when individual systems have heterogeneous error patterns \cite{jiang-etal-2023-llmblender,yang2023one,ai2025beyond}, we evaluate multiple ensembling techniques to combine the four prompting approaches. While ensembles improve average performance, we still observe patterns where a single prompting method remains preferable for particular personas or base models. To address this, we propose a lightweight meta-ensemble that treats the four prompting outputs as a binary prediction vector and learns a discriminative combiner using a support vector machine (SVM) \cite{cortes-vapnik-1995-svm}. Across our experiments, this SVM-based ensemble achieves the \textbf{best} overall performance and is the \textbf{only} approach that consistently outperforms each individual prompting strategy.

In sum, our contributions are (i) carrying out one of the first systematic comparisons of persona-conditioned prompting methods for toxicity detection for different personas, showing that no single method is superior; 
(ii) introducing a prompt-optimization approach for persona prompting based on TextGrad \cite{yuksekgonul-etal-2024-textgrad} with performance comparable to value profiles \cite{sorensen-etal-2025-value}; (iii) benchmarking a range of ensembling strategies inspired by prior LLM aggregation work \cite{jiang-etal-2023-llmblender,yang2023one,ai2025beyond}; and (iv) proposing an SVM meta-ensemble over binary prompt outputs that achieves the strongest results overall \cite{cortes-vapnik-1995-svm}.

\section{Related Work}
In this section, we review offensive language detection and biases of LLMs. Then, we review work on subjectivity, concluding the section with a discussion of work on ensembling LLMs.

\paragraph{Toxicity and offensive language detection.}
Some work study the automatic detection of toxic, hateful, or offensive language in online text, often using labeled data built from social media and forum data.
Early and widely used resources include Twitter-based hate/offensive \cite{davidson2017automated,founta2018large,wulczyn2017exmachinapersonalattacks} data.
Some tasks, such as OffensEval, further attempt to standardize evaluation for offensive language identification \citep{zampieri2019semeval}.
More recent datasets expand coverage and difficulty by adding spans \citep{mathew2021hatexplain}, targeting subtle/implicit hate \citep{elsherief-etal-2021-latent}, or generating large-scale synthetic toxic/benign examples about minority groups to test spurious correlations \citep{hartvigsen-etal-2022-toxigen}. LLMs have been also explored as automated classifiers for hateful/offensive/toxic content, often achieving reasonably strong agreement with crowd labels but remaining sensitive to prompt wording \cite{li2023hot,mishra2023githubtox}. Industry product deployments similarly report using LLMs to support moderation decisions, motivating research on reliability and bias, when LLM judgments substitute for or guide human moderation workflows \cite{markov2022holistic,openai2023gpt4moderation}. Most of these work aggregate labels into a single one, ignoring the importance of background and culture for toxicity perception.

\paragraph{Bias and unintended correlations in toxicity models.} Machine learning models might rely on surface patterns and have biases towards some texts, such as swear words. Dialect variation further amplifies this issue: demographic-aligned language features (e.g., African-American English) differ from mainstream norms \citep{blodgett-etal-2016-demographic}, and insensitivity to dialect has been shown to induce racial bias in hate-speech annotation and downstream classifiers \citep{sap-etal-2019-risk}. These findings motivate approaches that explicitly account for contextual and social perspectives when judging offensiveness, rather than treating toxicity as a purely lexical phenomenon.

\paragraph{Subjectivity, annotator effects, and pluralistic labeling.}
Toxicity perception is inherently subjective and shaped by context, norms, and background. More generally, annotation research argues that disagreements can be informative rather than noise \citep{aroyowelty2015crowdtruth,fleisig-etal-2023-majority}.
Systematic reviews of abusive-language datasets highlight that annotation guidelines, sampling strategies, and annotator populations vary widely, affecting both model performance and fairness \citep{vidgen2020directions}.
This line of work suggests evaluation setups should consider multiple perspectives and analyze disagreement patterns instead of aggregating labels into a single ground truth.

\begin{table*}[th!]
\begin{center}
\small
\begin{tabular}{l|p{10cm}}\hline 
\textbf{Prompt Type} & \textbf{Prompt} \\ \hline
Definition &Offensiveness denotes the overall rudeness, disrespect, or toxicity of a post. whether a post could be considered offensive to anyone. \\\hline
Persona  & You are supposed to decide if the given text toxic or offensive based on the following definition:
\textit{\{definition\}}
Imagine you were {persona} and how you would feel about the given statement if you were \textit{\{persona\}}. \\\hline
Value Profile &  You are supposed to decide if the given text toxic or offensive based on the following definition:
\textit{\{definition\}}
Value Profile: \textit{\{Value Profile\}}
\\\hline

Default & You are an helpful assistant and expert in linguistic analysis. You will decide if the given statement is offensive. 
\textit{\{definition\}}  \\\hline

\end{tabular}
\end{center}
\caption{Prompts used in baseline methods and the definition of toxicity used in this work in the prompts.}
\label{tab:persona_prompts} 
\end{table*}

\paragraph{Aggregating multiple LLM judgments.}
Because LLM outputs can be non-deterministic and sensitive to prompting, aggregation can improve robustness.
Self-consistency aggregates multiple reasoning traces via sampling and voting to increase accuracy \citep{wang2022selfconsistency}.
Multi-prompt ensembling works explore weighted voting and dynamic selection to exploit complementary strengths across systems \citep{yang2023one,ai2025beyond}.
In toxicity detection settings where labels are subjective, aggregation can also be viewed as approximating a mixture of perspectives, making it especially relevant for analyzing disagreement and bias.

Building on research in biased toxicity detection \citep{dixon2018unintended,borkan2019nuanced,sap2019racialbias} and pluralistic annotation \citep{aroyowelty2015crowdtruth,waseem2016annotator},
we study how demographic perspectives can be operationalized via prompting for offensiveness classification and their influence on both overall performance and subgroup behavior.
We further connect this analysis to LLM aggregation methods \citep{ai2025beyond,yang2023one} to evaluate whether combining judgments improves reliability. We believe this is one of the first systematic work that explores LLMs as judges for toxicity detection for different demographic perspectives.

\section{Methods}
We experiment with different prompts that help guide LLMs to think from the perspective of a persona. We also have a ``default'' prompt that does not assign any persona. In addition to these, we experiment with ensembling techniques to combine the predictions from the same model with different prompts. Lastly, we propose a prompting method and an ensembling method, which, in combination, perform better than previous methods.

\subsection{Prompting Methods}First, we examine the effect of assigning a persona to an LLM. We experiment with perspective-taking prompts \cite{xu-etal-2024-walking}, where we instruct the model to condition its toxicity judgments on the specified persona. 
Further, we experiment with Value Profiles \cite{sorensen2025value}, which are natural language descriptions of underlying values extracted from in-context examples. The aim is to preserve the information from in-context examples, but in a natural and explicit description. 
Similar to \citet{sorensen2025value}, we used Gemini1.5 Pro to create value profiles. For each persona, we selected 7 offensive and 7 non-offensive examples from the examples where there are disagreements among different annotators.  Each prompt can be found in Table \ref{tab:persona_prompts}. 

\subsection{Ensembling Techniques} Majority voting is common to combine predictions from different models \cite{ai2025beyond,yang2023one} or humans \cite{galton1907vox}. 
Hence, we explore weighted and unweighted majority voting approaches and propose a simple ensembling method. Here, we combine different prompting methods for the same model instead of combining predictions from different models. Because all of the ensembling approaches require training data, we reserve 20\% of the data for train, 10\% for validation, and 70\% for the test for each persona. All results including single prompting techniques are reported on the same test set.  

The first ensembling approach we experiment with is \citet{yang2023one} which is a weighted majority voting technique where the weights are correlated with the accuracy of the approach on a ``train'' data. We call this approach \textbf{``Accuracy-based Weighted Majority voting''}. Second, we apply the weights proposed by \citet{ai2025beyond}, the ``optimal weight'' for a model ``m'' is $log_e{\frac{acc_m}{1-acc_m}}$. We call this approach \textbf{``Theoretical Optimal Weighted Majority Voting''}. This is similar to \citet{yang2023one} in terms of adjusting weights based on accuracy but they use a different formula. Further, as a baseline we have the \textbf{Best Unweighted Majority Voting}, which looks at all possible subsets of the prompting techniques (default, persona, optimized persona, and value profile) and chooses the best performing one on the test set (so this approach can be considered as the oracle approach which has the hypothetical maximum accuracy you can get with an unweighted ensembling method). This approach gets the best possible accuracy on the test set when an unweighted majority voting is used. When there is a tie, we choose to be conservative so if there is at least one vote for offensive, we aggregate them as offensive. 

\subsection{Proposed Prompting and Ensembling Approaches}
Inspired by \citet{sorensen2025value} and the potential of automated prompt optimization \cite{yuksekgonul2024textgrad}, we suggest optimizing persona prompts using automated prompt engineering (APE) techniques for each persona and model so we can force the models to think from the perspective of the corresponding persona effectively. This is a prompting-based approach and thanks to optimization we aim to have the most effective prompts for each persona and model. We use TextGrad \cite{yuksekgonul2025optimizing} as an APE which uses an additional LLM (the 'optimizer') 
that criticizes and suggests new prompts based on the loss from the current prompt. We sample 100 examples for training and 100 for validation, our loss function is accuracy, we and use Gpt4o as an optimizer LLM. To the best of our knowledge, no work explores APE for toxicity detection, especially for representing the perspectives of different personas.

Previous ensembling approaches rely mostly on the accuracy of the methods on a training set to adjust the weights of the methods. Hence, all of them are \textbf{linear} combiners. However, non-linear methods might have a higher potential. On the other hand, the ensembling process already has computational costs. Based on these factors, we propose a simple but efficient classifier that takes a binary vector of 4 representing predictions from different prompting methods (default, persona, value profile, optimized persona) for each persona. The classifier of our ensemble is \textbf{SVM} with a Gaussian Kernel to make the function non-linear. As train and validation data, we use the same splits that we use for the other ensembling approaches.
\section{Dataset and Models}

\subsection{Dataset}
We experiment with the Social Bias Frames dataset \cite{sap2020social}, which has 44k unique social media posts with annotations for offensiveness, target group etc. Also, for each post, they have multiple annotations with demographic information of the annotator. In this work, we focus on the combination of gender and race (we refer to each gender and race combination as a ``persona.''). We choose the personas that have a reasonable ($>400$) number of examples. Table \ref{dataset_statistics} gives the statistics about the dataset for each persona.

\begin{table}[t!]
\begin{center}
\footnotesize
\begin{tabular}{l|r|r}\hline 
\multicolumn{1}{c|}{\textbf{Persona}} & 
\multicolumn{1}{c|}{\textbf{\# of Examples}} & 
\multicolumn{1}{c}{\textbf{Offensiveness \%}}  \\\hline
Asian Woman & 2829 & 51.11 \\
Asian Man & 4468 & 36.59 \\
Black Woman & 4414 & 54.05 \\
White Woman & 32125 & 54.09 \\
White Man & 31964 & 56.65 \\
Hispanic Woman & 4109 & 65.40 \\
Hispanic Man & 3007 & 50.05 \\
Native Man & 471 & 40.51 \\
\hline
\end{tabular}
\end{center}
\caption{Dataset Statistics with number of examples and percentage of offensiveness labels.}
\label{dataset_statistics} 
\end{table}

\begin{figure}[th!]
  \centering
  \includegraphics[width=\linewidth]{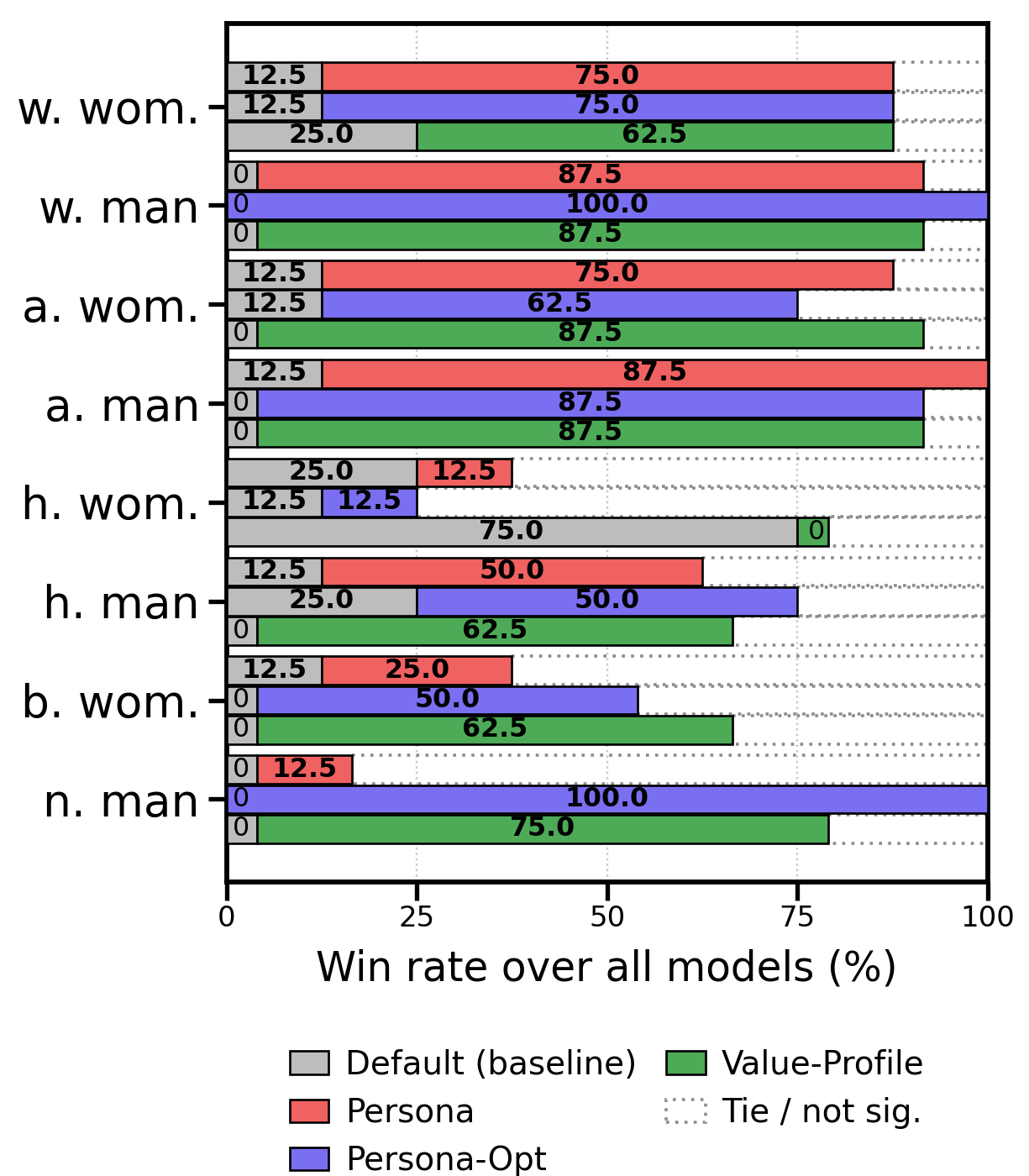}
  \caption{Prompting comparisons with \textbf{default prompting} as the baseline.}
  \label{fig:prompt_baseline_default}
\end{figure}



\subsection{Models}
We experiment with Llama3.1 70b/8b \cite{dubey2024llama}, and Qwen2.5 14b/32b \cite{qwen2025qwen25technicalreport} models. Additionally, we analyze the effect of reasoning-enhanced models. For this, we use R1-distilled models (e.g. R1 Distilled Qwen 14b/32b and R1 Distilled Llama 8b/70b) \cite{guo2025deepseek}.
\section{Results}
In all results, we apply McNemar's test \cite{McNemar1947} to compare approaches pairwise and report percentage wins of each method against each other. The McNemar is a paired, nonparametric test for two related binary data, e.g., two classifiers on the same data, and checks whether their error rates differ significantly. We analyze the effects of reasoning models, persona prompts, and ensembling methods. We conclude this section with an overall comparison and an ablation study of the SVM.

\subsection{Reasoning Effect}
First, we compare the default prompts for the base models and their reasoning-enhanced counterparts. We see that reasoning helps for smaller models but degrades the performance for larger models (See the top portion of Table \ref{tab:all_results} in Appendix). This might indicate that large models are already doing some reasoning internally and enhancing their reasoning mostly for logical and mathematical tasks hurts their social reasoning. Also, we apply McNemar's test to compare if the model predictions are different for each persona, and in most cases they are significantly different.

\subsection{Persona-based Prompt Effects}

Figure \ref{fig:prompt_baseline_default} shows the comparison of prompts against default (no persona) prompt aggregated over models (see Table \ref{tab:base_default} in Appendix for all results). Overall, all three persona-based methods outperform the default prompt with varying success based on the models and persona. They work better for non-reasoning models which might indicate that reasoning models might implicitly think from the perspective of some personas, but non-reasoning models need a trigger to do that. None of these works well for ``hispanic woman,'' and value profile makes the predictions worse. In our dataset, hispanic woman is the persona that has the most offensive content. This motivated us to check the predictions and our analysis shows that when a persona prompt is used, models tend to predict non-offensive more often.

\begin{figure}[th!]
  \centering
  \includegraphics[width=\linewidth]{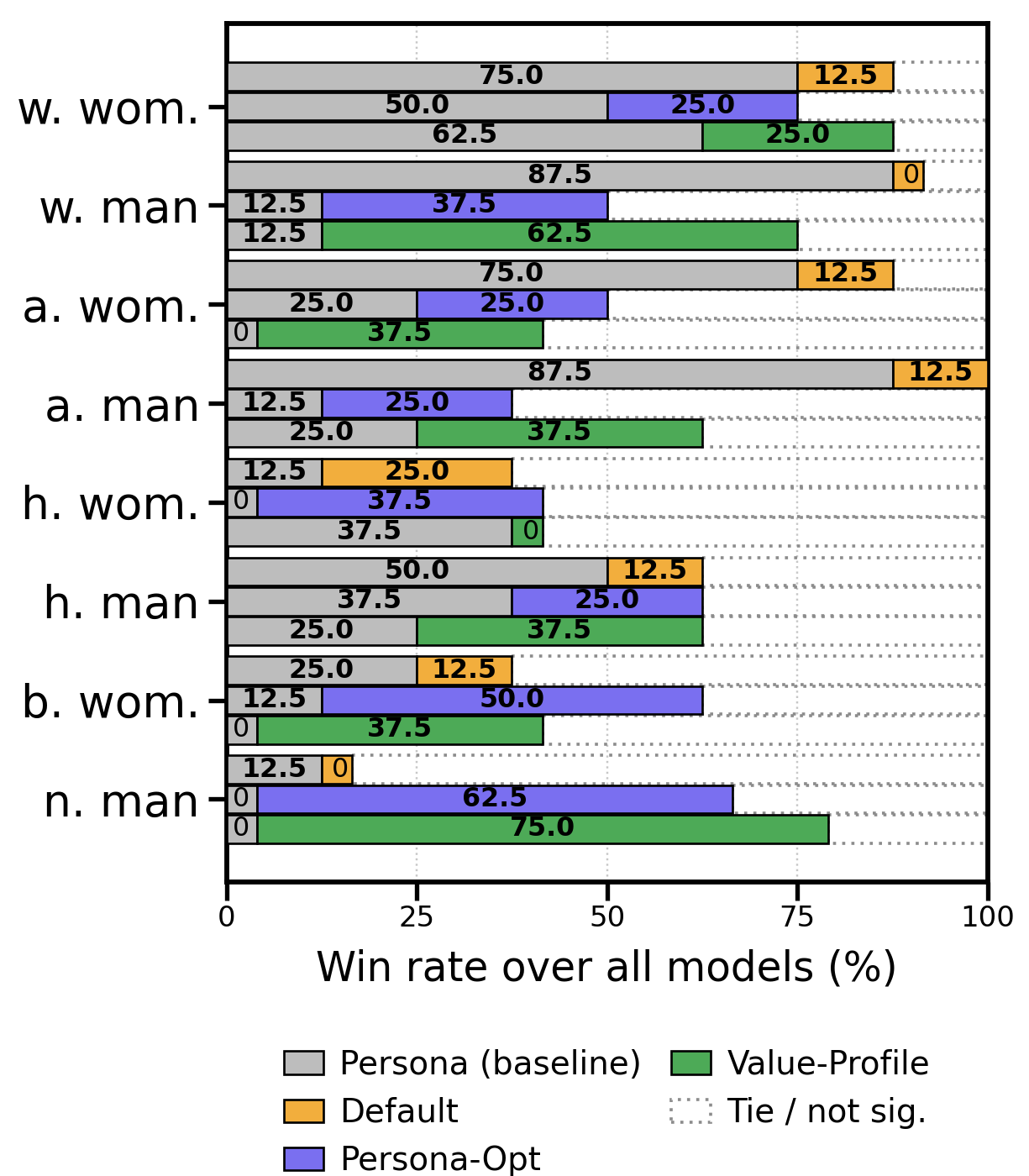}
  \caption{Prompting comparisons with \textbf{persona prompt} as the baseline.}
  \label{fig:prompt_baseline_persona}
\end{figure}

\paragraph{Persona Prompt}
Simply assigning persona information to reflect the offensive content perception of that persona usually improves the performance (e.g. compare gray and pink boxes in Figure \ref{fig:prompt_baseline_default}). It works for most models except Distilled Llama8b and works better for non-reasoning models (see Table \ref{tab:base_default}).

Regarding the effect on different personas, Figure \ref{fig:prompt_baseline_default} shows that persona prompt degrades performance for Hispanic woman and does not have much effect for Native man. It also does not improve results much for Black woman. For the other personas, it boosts performance.

\begin{figure}[th!]
  \centering
  \includegraphics[width=\linewidth]{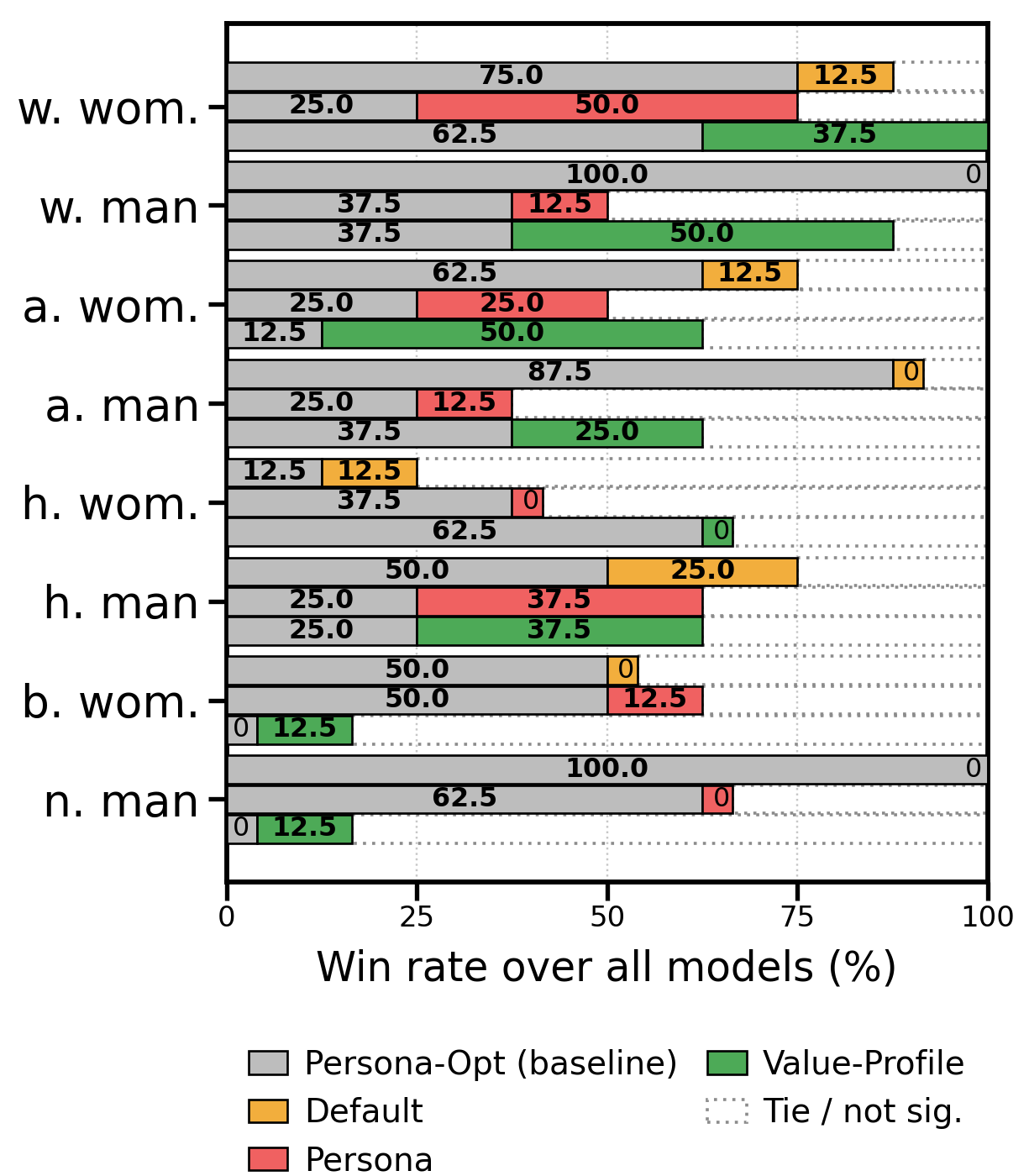}
  \caption{Prompting comparisons with \textbf{persona optimized prompt} as the baseline.}
  \label{fig:prompt_baseline_personaopt}
\end{figure}

\paragraph{Persona optimized Prompt}
Similarly, when the persona prompt is optimized for each model and persona, it mostly boosts performance (in Figure \ref{fig:prompt_baseline_default} purple boxes outperform gray ones.). It also boosts performance for reasoning models; performance improvement for non-reasoning models is still greater, though (see Table \ref{tab:base_default} in Appendix). Optimization improves performance on all personas except Hispanic woman.

Optimizing persona prompt outperforms the persona prompt overall. (see Figure \ref{fig:prompt_baseline_persona} and  Table \ref{tab:base_persona} in Appendix). However, it does not improve much for Llama70b, Distilled Llama70b, Qwen14b/32b and Distilled Qwen32b models. Further, it works better for White man, Hispanic woman, Black woman, and Native man. For the other personas, both approaches are comparable.

\paragraph{Value Profile Prompt}
Figure \ref{fig:prompt_baseline_default} shows that providing value profiles outperforms the default prompt significantly in most cases. Its effect is similar to optimizing prompts, and it improves performance for reasoning models as well, except for Distilled Qwen14b (c.f Table \ref{tab:base_default} in Appendix).  Further, it improves performance on all personas except Hispanic woman.

Providing value profiles is also a better approach than persona prompt (see Figure \ref{fig:prompt_baseline_persona} and  Table \ref{tab:base_persona} in Appendix). However, the performances are comparable for Qwen14b, Distilled Qwen14b and Qwen32b models. Value profile is more effective, especially for Llama-based models. Regarding personas, value profile is better for White man, Black woman, and Native man. They are comparable for Asian man and Hispanic woman, but the persona prompt is more effective for White woman and Hispanic woman.

Based on Figure \ref{fig:prompt_baseline_personaopt}, we can see that value profile prompting, and optimizing prompts for each model and persona perform similarly (see Table \ref{tab:base_val_prof} for more details). One of the methods works better for some models and the other one works better for the others. We have a similar observation for personas as well. The reason behind this might be that value profile prompting also tailors prompts for each persona. Therefore, it somehow also ``optimizes'' the prompt in a more guided way.

Overall, the ordering of the methods is \textbf{Optimized Prompt = Value Profile > Persona Prompt > default Prompt}

\begin{figure}[th!]
  \centering
  \includegraphics[width=\linewidth]{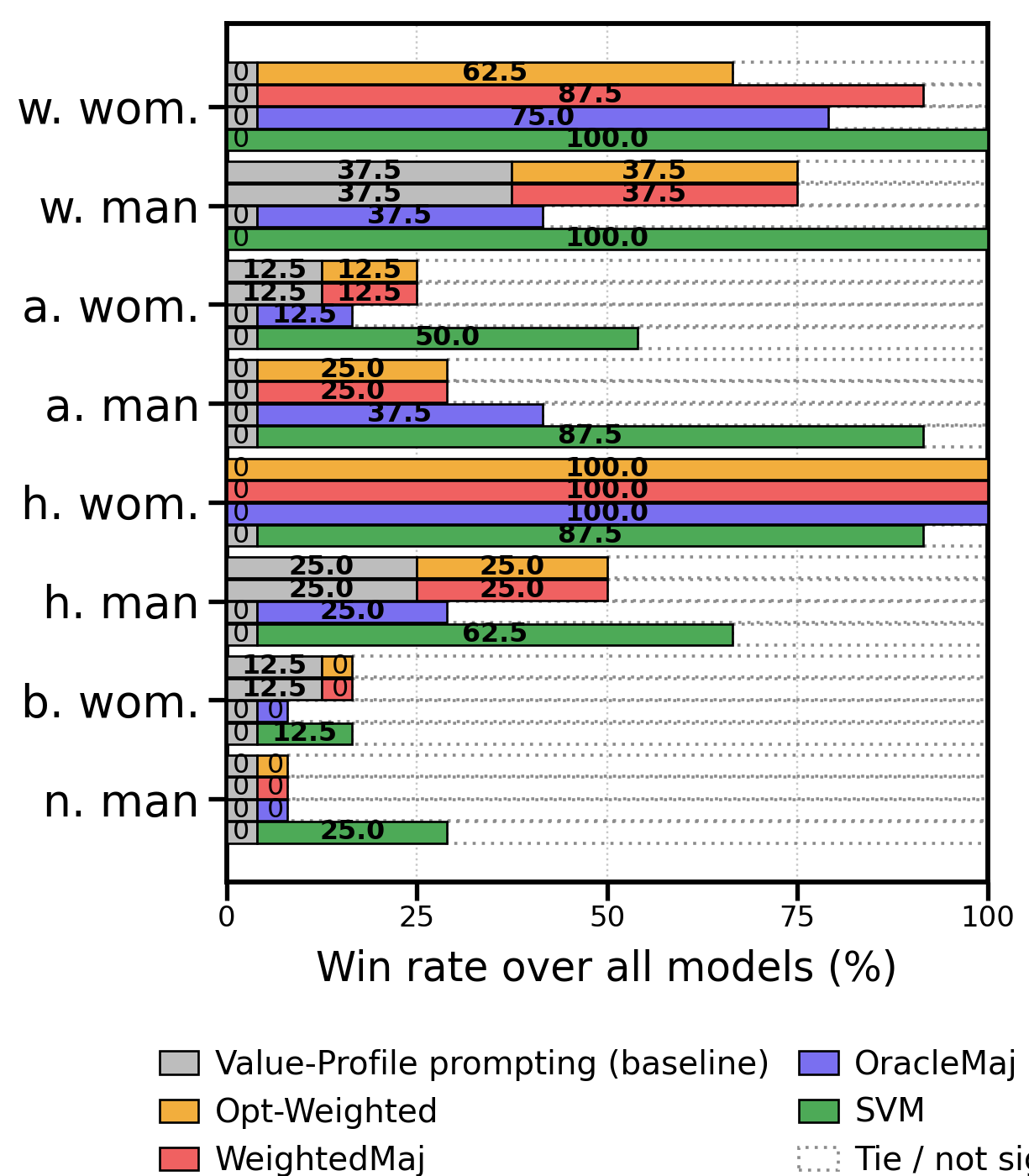}
  \caption{Ensembling comparisons with \textbf{Value Profile prompting} as the baseline.}
  \label{fig:ensembling_baseline_valprof}
\end{figure}

\subsection{Ensembling Methods}
Here, we combine the predictions of previous prompting methods for each model in different ways and compare the effect of ensembling methods.


Figure \ref{fig:ensembling_baseline_valprof} (see Table \ref{tab:base_val_prof} in Appendix more details) compares ensembling approaches against value profile prompting which is one of the best performing prompting method. All ensembling methods outperform it in most cases. Value profile prompt is better than Accuracy-based Weighted Majority Voting for Distilled Llama8b, Asian woman, and Black woman. They are comparable for Distilled Qwen32b, White man, and Native man. Best Unweighted Majority Voting is similar to value profile for Distilled Llama8b, Distilled Qwen32b, Asian woman, Black woman, and Native man. Theoretical Optimal Weighted Majority Voting performs worse than value profile for Distilled Llama8b, Distilled Qwen32b Asian woman, and Black woman. They perform similarly for White man and Native man. For the comparison to other prompting methods see Tables \ref{tab:base_default}, \ref{tab:base_persona}, \ref{tab:base_persona_opt} in Appendix.


\begin{figure}[t]
  \centering
  \includegraphics[width=\linewidth]{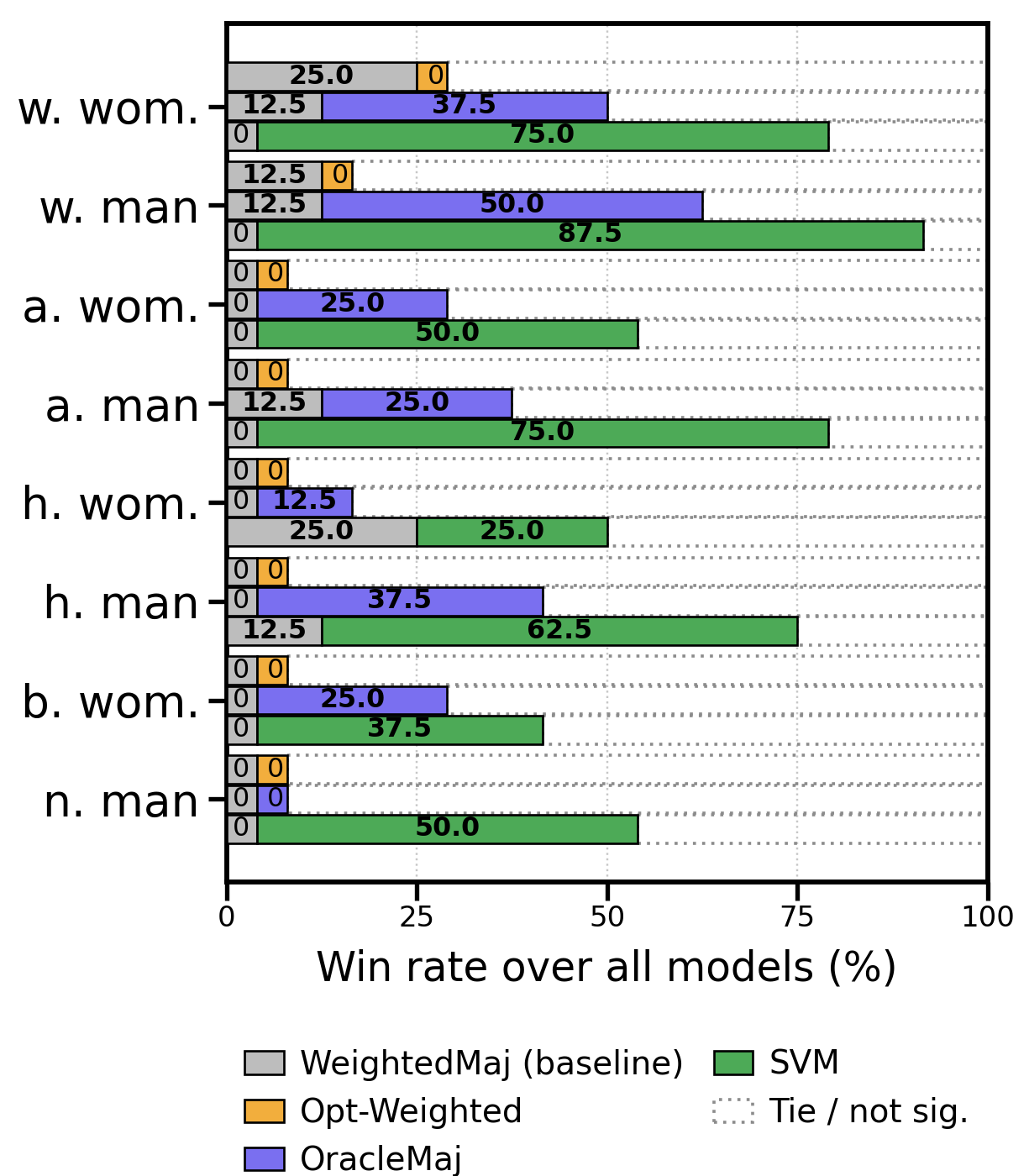}
  \caption{Ensembling comparisons with \textbf{Accuracy-based Weighted Majority} as the baseline.}
  \label{fig:ensembling_baseline_weightedmaj}
\end{figure}

\begin{figure}[t]
  \centering
  \includegraphics[width=\linewidth]{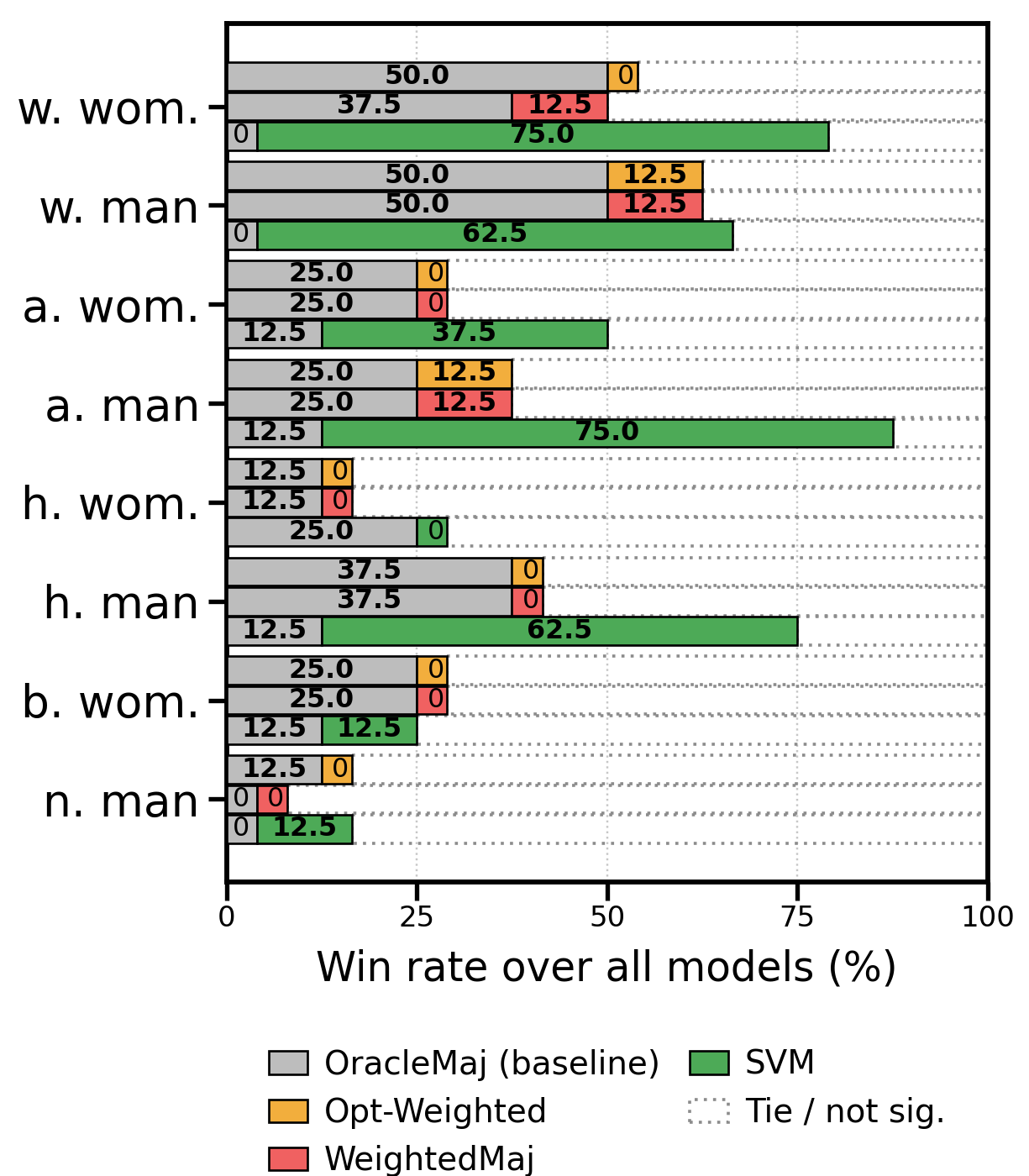}
  \caption{Ensembling comparisons with \textbf{Best Unweighted Majority} as the baseline.}
  \label{fig:ensembling_baseline_oraclemaj}
\end{figure}

\begin{figure}[t]
  \centering
  \includegraphics[width=\linewidth]{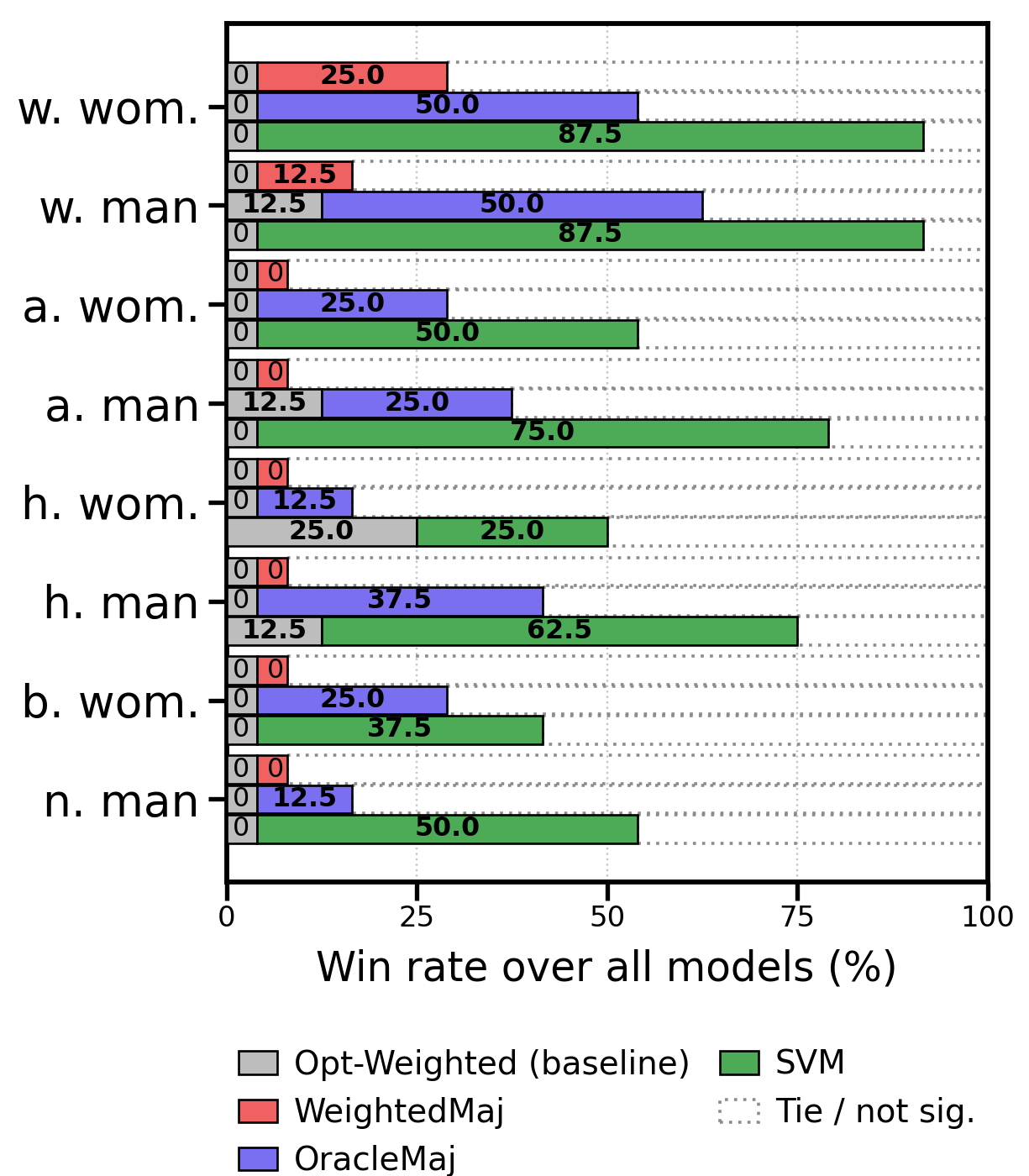}
  \caption{Ensembling comparisons with \textbf{Theoretical Optimal Weighted Majority} as the baseline.}
  \label{fig:ensembling_baseline_opt_weighted}
\end{figure}


\paragraph{Accuracy-based Weighted Majority Voting}
This ensembling method outperforms each single prompting method in most cases. Its effectiveness changes depending on a model and persona.

\paragraph{Best Unweighted Majority Voting}
Since the purpose of this approach is to set an upper limit for an unweighted majority voting approaches, it improves performance against persona prompting for all models and personas as expected. There are a few cases, where both approaches are comparable. We should note that we choose the best combination on test set, hence this method has an advantage over the other methods.

This ensembling method is better than Accuracy-based Weighted Majority Voting (see Figure \ref{fig:ensembling_baseline_weightedmaj} and Table \ref{tab:base_weighted_maj_voting} in Appendix), though its effectiveness differ based on the model and persona. They are comparable for Distilled Qwen14b/32b, Native man and Hispanic woman.

\paragraph{Theoretical Optimal Weighted Majority Voting}
This performs similar to Accuracy-based Weighted Majority Voting so it also outperform single prompting methods in most cases; though, its performance vary based on a model and persona. These two methods are comparable, none of them is superior. On the other hand, this method is worse than Best Unweighted Majority Voting in most cases (see Figure \ref{fig:ensembling_baseline_oraclemaj} and Table \ref{tab:base_best_maj_voting}).

\paragraph{SVM}
SVM trained on a binary vector of predictions from single prompting methods is the most consistent and powerful approach. It outperforms single prompting methods in \textbf{all} cases. Its effectiveness changes based on a model and persona but it is always better.

When we compare this method against Accuracy-based Weighted Majority Voting (Figure \ref{fig:ensembling_baseline_weightedmaj} and Table \ref{tab:base_weighted_maj_voting} in Appendix) and Theoretical Optimal Weighted Majority Voting (Figure \ref{fig:ensembling_baseline_opt_weighted}and Table \ref{tab:base_theo_opt} in Appendix), we see that SVM outperforms both methods in most cases. SVM is comparable to both methods for Llama70b.

Figure \ref{fig:ensembling_baseline_oraclemaj} shows that SVM is significantly better than Best Unweighted Majority Voting too (more details are in Table \ref{tab:base_best_maj_voting} in Appendix). They are comparable for Llama8b/70b and Hispanic woman. This shows that weighted ensembling methods have a better potential than unweighted ensembling methods for hate speech detection based on personas. Even an SVM classifier with an input of binary vector of predictions for different promptings can learn to predict better for a given persona.

Overall, the ordering of the methods is \textbf{SVM > Best Unweighted Majority Voting > Theoretical Optimal Weighted Majority Voting = Accuracy-based Weighted Majority Voting}. SVM is the only method including other ensembling methods and single prompting techniques that is consistently better for all models and personas.

\begin{table*}[h!]
\begin{center}
\footnotesize
\begin{tabular}{lrrrrrrr}\hline 
\multicolumn{1}{c}{} & 
\multicolumn{1}{c}{\textbf{Pers. Prompt}} & 
\multicolumn{1}{c}{\textbf{Pers. opt}} & 
\multicolumn{1}{c}{\textbf{Value Prof.}} & 
\multicolumn{1}{c}{\textbf{Weigh. Maj.}} & 
\multicolumn{1}{c}{\textbf{Best Maj.}} & 
\multicolumn{1}{c}{\textbf{Weigh. Maj. theo.}} & 
\multicolumn{1}{c}{\textbf{SVM}}  \\

Default & 7 (34) & 5 (43) & 8 (42) & 0 (50) & 0 (57) & 0 (50) & 0 (58) \\
Pers. Prompt & & 12 (23) & 13 (25) & 4 (37) & 0 (40)& 4 (36) & 1 (51) \\
Value Prof & & & 18 (19) & 10 (25) & 0 (24)& 10 (23) & 0 (47) \\
Pers. opt & & &  & 4 (23) & 0 (28) & 6 (23) & 1 (44) \\
Weigh. Maj. & & & & & 2 (20)& 3 (0) & 3 (40) \\
Best Maj. & & & & && 21 (1) & 2 (29) \\
Weigh. Maj. theo. & & & & && & 3 (41)\\\hline

\end{tabular}
\end{center}

\caption{Out of 64 combinations (8 models x 8 personas) how many times the method on y axis is better than the method on x axis (the method on x is better than the method on y axis)}
\label{tab:pairwise_comp_all} 
\end{table*}

\subsection{Overall Comparison}
Table \ref{tab:pairwise_comp_all} compares the significant differences among each method pairwise and Table \ref{tab:all_results} in Appendix compares each method in terms of F1 scores. They show that ensembling methods outperform single prompting methods and SVM is the best method overall.  In some cases, Best Unweighted Majority Voting outperforms the others, but in most cases, they are not significant and that method has an advantage over other methods by choosing the best combination on test set.

\begin{figure}[t]
  \centering
  \includegraphics[width=\linewidth]{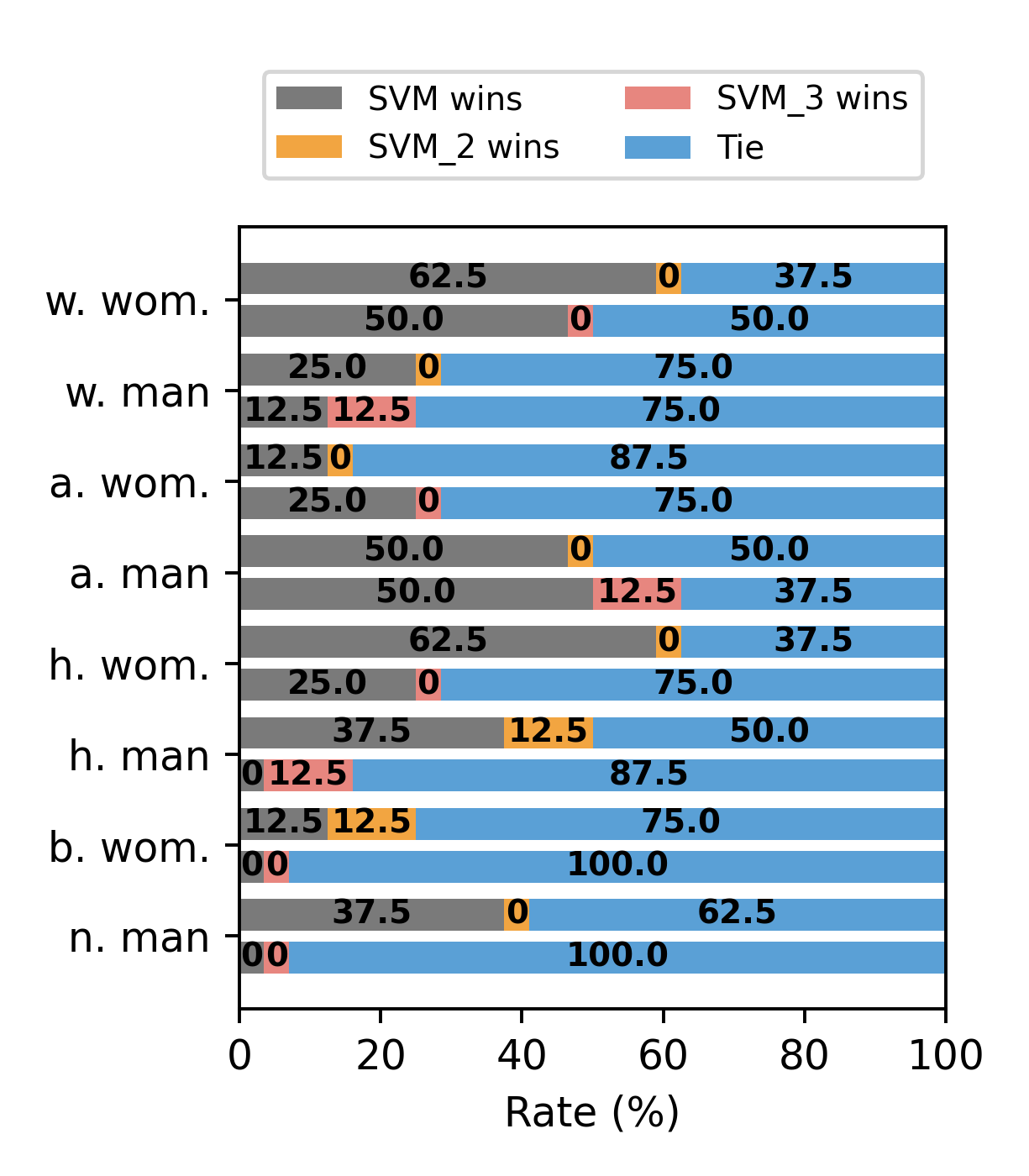}
  \caption{Significant test comparison of ablation study. SVM is the original model, SVM\_2 uses two inputs and SVM\_3 uses three different inputs.}
  \label{fig:svm_ablation}
\end{figure}

\subsection{Ablation Study on SVM}
To see the effect of using different numbers of prompts, we conduct an ablation study where we vary the number of prompts as inputs to SVM. Our original SVM model takes predictions from all four prompting methods and the ablation models take either two or three. For each LLM, we pick the best combination of two and three promptings by looking at the accuracies of SVM trained on each combination. Figure \ref{fig:svm_ablation} compares the original SVM model with the best SVMs that use two or three inputs. These results show that the combination of the four prompts is superior to any subset, which suggests that they have complementary strengths unmatched by any pair or triple.

\section{Conclusion}

Toxicity detection is not only a modeling problem but also a perspective problem: demographically distinct personas 
can legitimately disagree on the toxicity of the same post. In this work, we evaluated persona-aware prompting strategies for toxicity detection and showed that while a persona conditioned prompt usually improves over a non-persona prompt, no single prompting method consistently dominates across all personas and base models. We show that reasoning-enhanced models do not inherently solve the challenge of subjective perception and can even degrade performance in larger models for social reasoning tasks. We further introduced an automated prompt-optimization strategy based on TextGrad that produces persona prompts competitive with value-profile conditioning, yet with disagreement patterns that indicate complementary strengths. To make use of this heterogeneity, we explored ensembling techniques and propose  a meta-ensembling approach using an SVM classifier over binary prompt outputs. This simple method is the only method that is consistently better across the full set of model--persona configurations. Our findings suggest that the future of pluralistic toxicity detection lies in learned ensembling of diverse prompting perspectives, providing a more reliable approach to model the subjective judgments of heterogeneous groups.
\section*{Limitations}
Our study focuses on limited set of personas. We do not look at queer gender types and some other common races. Further, we experiment with models from two family. Our findings might differ in closed-source models. In addition, we focus only on English but there are other languages that the models behave differently. Last but not least, performance disparities among different personas remain in all methods.


\bibliography{custom}

\appendix

\newpage
\section{Pairwise Comparison Tables}
  \begin{table*}[h!]
\begin{center}
\footnotesize
\begin{tabular}{l|r|r|r|r|r|r|r}\hline 
\multicolumn{1}{c|}{\textbf{Model/Persona}} & 
\multicolumn{1}{c|}{\textbf{Pers. Prompt}} & 
\multicolumn{1}{c|}{\textbf{Pers. opt}} & 
\multicolumn{1}{c|}{\textbf{Value Prof.}} & 
\multicolumn{1}{c|}{\textbf{Weigh. Maj.}} & 
\multicolumn{1}{c|}{\textbf{Best Maj.}} & 
\multicolumn{1}{c|}{\textbf{Weigh. Maj. theo.}} & 
\multicolumn{1}{c}{\textbf{SVM}}  \\\hline

llama8b & 7 (0) & 8 (0) & 7 (0) & 8 (0)  & 8 (0) & 8 (0) & 8 (0) \\
dllama8b & 0 (5) & 4 (1) & 4 (1) & 3 (0)  & 5 (0) & 3 (0) & 8 (0) \\
llama70b & 4 (1) & 5 (1) & 3 (1) & 5 (0)  & 8 (0) & 5 (0) & 6 (0) \\
dllama70b & 4 (0) & 4 (1) & 6 (2) & 7 (0)  & 7 (0) & 7 (0) & 7 (0) \\
qwen14b & 5 (0) & 6 (1) & 6 (1) & 6(0)  & 7 (0) & 7 (0) & 8 (0) \\
dqwen14b & 2 (0) & 4 (1) & 2 (2) & 7 (0)  & 7 (0) & 7 (0) & 7 (0) \\
qwen32b & 7 (1) & 6 (0) & 7 (0) & 7 (0)  & 7 (0) & 7 (0) & 7 (0) \\
dqwen32b & 5 (0) & 6 (0) & 6 (0) & 7 (0)  & 7 (0) & 7 (0) & 7 (0) \\
\textbf{Total} & 34 (7) & 43 (5) & 42 (8) & 50 (0)  & 57 (0) & 50 (0) & 58 (0) \\
\hline 
white woman & 6 (1) & 6 (1) & 5 (2) & 7 (0)  & 7 (0) & 7 (0) & 8 (0) \\
white man & 7 (0) & 8 (0) & 7 (0) & 8 (0)  & 8 (0) & 8 (0) & 8 (0) \\
asian woman & 6 (1) & 5 (1) & 7 (0) & 7 (0)  & 8 (0) & 8 (0) & 8 (0) \\
asian man & 7 (1) & 7 (0) & 7 (0) & 8 (0)  & 8 (0) & 8 (0) & 8 (0) \\
hispanic woman & 1 (2) & 1 (1) & 0 (6) & 3 (0)  & 3 (0) & 5 (0) & 3 (0) \\
hispanic man & 4 (1) & 4 (2) & 5 (0) & 7 (0)  & 7 (0) & 7 (0) & 8 (0) \\
black woman & 2 (1) & 4 (0) & 5 (0) & 4 (0)  & 4 (0) & 6 (0) & 8 (0) \\
native man & 1 (0) & 8 (0) & 6 (0) & 6 (0)  & 6 (0) & 8 (0) & 7 (0) \\\hline

\end{tabular}
\end{center}

\caption{Out of 8 models or personas, how many times is the approach significantly better than default prompt (how many times it is worse) Pers. Prompt refers to persona prompt, Pers. opt refers to Persona optimized prompt, Value Prof. refers to Value profile prompting, Weight. Maj. refers to Accuracy-based Weighted Majority Voting, Best Maj. refers to Best Unweighted Majority Voting, Weigh. Maj. theo. refers to Theoretical Optimal Weighted Majority Voting}
\label{tab:base_default} 
\end{table*}

\begin{table*}[t]
\begin{center}
\footnotesize
\begin{tabular}{l|r|r|r|r|r|r}\hline 
\multicolumn{1}{c|}{\textbf{Model/Persona}} & 
\multicolumn{1}{c|}{\textbf{Pers. opt}} & 
\multicolumn{1}{c|}{\textbf{Value Prof.}} & 
\multicolumn{1}{c|}{\textbf{Weigh. Maj.}} & 
\multicolumn{1}{c|}{\textbf{Best Maj.}} & 
\multicolumn{1}{c|}{\textbf{Weigh. Maj. theo.}} & 
\multicolumn{1}{c}{\textbf{SVM}}  \\\hline

llama8b  & 7 (0) & 6 (1) & 8 (0)  & 8 (0) & 8 (0) & 7 (0) \\
dllama8b  & 5 (1) & 7 (1) & 7 (0)  & 7 (0) & 7 (0) & 7 (0) \\
llama70b & 2 (2) & 4 (1) & 2 (0)  & 3 (0) & 2 (0) & 3 (0) \\
dllama70b  & 2 (4) & 4 (1) & 5 (0)  & 5 (0) & 5 (0) & 7 (0) \\
qwen14b  & 0 (0) & 3 (2) & 2(2)  & 3 (0) & 2 (2) & 8 (0) \\
dqwen14b &  4 (1) & 2 (3) & 6 (0)  & 8 (0) & 6 (0) & 7 (0) \\
qwen32b  & 1 (2) & 1 (3) & 1 (2)  & 2 (0) & 1 (2) & 4 (1) \\
dqwen32b & 2 (2) & 2 (0) & 6 (0)  & 4 (0) & 5 (0) & 8 (0) \\
\textbf{Total} & 23 (12) & 25 (13) & 37 (4)  & 40 (0) & 36 (4) & 51 (1) \\
\hline 
white woman  & 2 (4) & 2 (5) & 4 (2)  & 4 (0) & 3 (2) & 6 (0) \\
white man & 3 (1) & 5 (1) & 6 (0)  & 7 (0) & 6 (0) & 8 (0) \\
asian woman &  2 (2) & 3 (0) & 4 (0)  & 4 (0) & 4 (0) & 6 (0) \\
asian man & 2 (1) & 3 (2) & 2 (0)  & 4 (0) & 2 (0) & 7 (0) \\
hispanic woman  & 3 (0) & 0 (3) & 6 (0)  & 5 (0) & 6 (0) & 5 (0) \\
hispanic man & 2 (3) & 3 (2) & 4 (1)  & 5 (0) & 4 (1) & 7 (1) \\
black woman &  4 (1) & 3 (0) & 5 (1)  & 5 (0) & 5 (1) & 6 (0) \\
native man &  5 (0) & 6 (0) & 6 (0)  & 6 (0) & 6 (0) & 6 (0) \\\hline
\end{tabular}
\end{center}

\caption{Out of 8 models or personas, how many times is the approach significantly better than persona prompt (how many times it is worse)}
\label{tab:base_persona} 
\end{table*}

\begin{table*}[t]
\begin{center}
\footnotesize
\begin{tabular}{l|r|r|r|r|r|r}\hline 
\multicolumn{1}{c|}{\textbf{Model/Persona}} & 
\multicolumn{1}{c|}{\textbf{Pers. opt}} & 
\multicolumn{1}{c|}{\textbf{Weigh. Maj.}} & 
\multicolumn{1}{c|}{\textbf{Best Maj.}} & 
\multicolumn{1}{c|}{\textbf{Weigh. Maj. theo.}} & 
\multicolumn{1}{c}{\textbf{SVM}}  \\\hline

llama8b  & 4 (0)  & 3 (1)  & 4 (0) & 2 (1) & 4 (0) \\
dllama8b  & 5 (1) & 1 (2)  & 1 (0) & 1 (2) & 7 (0) \\
llama70b & 5 (1)  & 5 (1)  & 4 (0) & 5 (1) & 5 (0) \\
dllama70b  & 1 (5)  & 2 (1)  & 2 (0) & 2 (1) & 5 (0) \\
qwen14b  & 1 (2)  & 3 (2)  & 3 (0) & 3 (2) & 7 (0) \\
dqwen14b &  4 (1)  & 5 (0)  & 5 (0) & 7 (0) & 7 (0) \\
qwen32b  & 3 (2)  & 4 (1)  & 4 (0) & 4 (1) & 5 (0) \\
dqwen32b & 0 (3)  & 2 (2)  & 1 (0) & 1 (2) & 7 (0) \\
\textbf{Total} & 19 (18)  & 25 (10)  & 24 (0) & 23 (10) & 47 (0) \\

\hline 
white woman  & 5 (3)  & 7 (0)  & 6 (0) & 5 (0) & 8 (0) \\
white man & 3 (4)  & 3 (3)  & 3 (0) & 3 (3) & 7 (0) \\
asian woman &  1 (4)  & 1 (3)  & 1 (0) & 1 (3) & 5 (0) \\
asian man & 3 (2)  & 3 (1)  & 3 (0) & 3 (1) & 8 (0) \\
hispanic woman  & 5 (0)  & 8 (0)  & 8 (0) & 8 (0) & 8 (0) \\
hispanic man & 2 (3)  & 3 (2)  & 3 (0) & 3 (2) & 6 (0) \\
black woman &  0 (1)  & 0 (1)  & 0 (0) & 0 (1) & 2 (0) \\
native man &  0 (1)  & 0 (0)  & 0 (0) & 0 (0) & 3 (0) \\\hline
\end{tabular}
\end{center}

\caption{Out of 8 models or personas, how many times is the approach significantly better than value profile prompt (how many times it is worse)}
\label{tab:base_val_prof} 
\end{table*}

\begin{table*}[t]
\begin{center}
\footnotesize
\begin{tabular}{l|r|r|r|r|r}\hline 
\multicolumn{1}{c|}{\textbf{Model/Persona}} & 
\multicolumn{1}{c|}{\textbf{Weigh. Maj.}} & 
\multicolumn{1}{c|}{\textbf{Best Maj.}} & 
\multicolumn{1}{c|}{\textbf{Weigh. Maj. theo.}} & 
\multicolumn{1}{c}{\textbf{SVM}}  \\\hline

llama8b  & 0 (2)  & 2 (0) & 0 (4) & 3 (0) \\
dllama8b  &  2 (0)  & 2 (0) & 2 (0) & 7 (0) \\
llama70b &  3 (0)  & 5 (0) & 3 (0) & 4 (0) \\
dllama70b  &  5 (1)  & 5 (0) & 4 (0) & 6 (0) \\
qwen14b  &  4 (1)  & 5 (0) & 4 (1) & 8 (0) \\
dqwen14b &  4 (0)  & 5 (0) & 4 (0) & 6 (0) \\
qwen32b  &  1 (0)  & 2 (0) & 1 (0) & 5 (1) \\
dqwen32b & 4 (0)  & 3 (0) & 4 (0) & 5 (0) \\
\textbf{Total}   & 23 (4)  & 28 (0) & 23 (6) & 44 (1) \\
\hline 
white woman  & 4 (2)  & 4 (0) & 4 (3) & 8 (0) \\
white man  & 5 (0)  & 7 (0) & 5 (1) & 8 (0) \\
asian woman &   3 (0)  & 4 (0) & 3 (0) & 6 (0) \\
asian man &  2 (1)  & 3 (0) & 2 (1) & 6 (0) \\
hispanic woman  & 2 (1)  & 3 (0) & 2 (1) & 2 (0) \\
hispanic man &  3 (0)  & 3 (0) & 3 (0) & 5 (1) \\
black woman &   3 (0)  & 3 (0) & 3 (0) & 5 (0) \\
native man &   1 (0)  & 2 (0) & 1 (0) & 4 (0) \\\hline
\end{tabular}
\end{center}

\caption{Out of 8 models or personas, how many times is the approach significantly better than persona optimized prompt (how many times it is worse)}
\label{tab:base_persona_opt} 
\end{table*}

\begin{table}[t]
\begin{center}
\footnotesize
\begin{tabular}{l|r|r|r}\hline 
\multicolumn{1}{c|}{\textbf{Model/Persona}} & 
\multicolumn{1}{c|}{\textbf{Best Maj.}} & 
\multicolumn{1}{c|}{\textbf{Weigh. Maj. theo.}} & 
\multicolumn{1}{c}{\textbf{SVM}}  \\\hline

llama8b  & 4 (0)  & 0 (2)  & 4 (0)  \\
dllama8b  & 2 (0) & 0 (0)  & 7 (0)  \\
llama70b & 3 (0)  & 0 (0)  & 2 (1)  \\
dllama70b  & 2 (0)  & 0 (0)  & 5 (0) \\
qwen14b  & 4 (0)  & 0 (0)  & 8 (0)  \\
dqwen14b &  1 (1)  & 0 (0)  & 3 (1)  \\
qwen32b  & 2 (0)  & 0 (0)  & 4 (1)  \\
dqwen32b & 2 (1)  & 0 (1)  & 7 (0)  \\
\textbf{Total} & 20 (2)  & 0 (3)  & 40 (3)  \\\hline 
white woman  & 3 (1)  & 0 (2)  & 6 (1)  \\
white man & 5 (1)  & 0 (1)  & 7 (0)  \\
asian woman &  3 (0)  & 0 (0)  & 5 (0)  \\
asian man & 3 (0)  & 0 (0)  & 7 (0) \\
hispanic woman  & 1 (0)  & 0 (0)  & 3 (1)  \\
hispanic man &  3 (0)  & 0 (0)  & 5 (1)  \\
black woman &  2 (0)  & 0 (0)  & 3 (0)  \\
native man &  0 (0)  & 0 (0)  & 4 (0)  \\\hline
\end{tabular}
\end{center}

\caption{Out of 8 models or personas, how many times is the approach significantly better than weighted majority voting (how many times it is worse). This compares ensembling methods}
\label{tab:base_weighted_maj_voting} 
\end{table}

\begin{table}[t]
\begin{center}
\footnotesize
\begin{tabular}{l|r|r|r}\hline 
\multicolumn{1}{c|}{\textbf{Model/Persona}} & 
\multicolumn{1}{c|}{\textbf{Weigh. Maj. theo.}} & 
\multicolumn{1}{c}{\textbf{SVM}}  \\\hline

llama8b  & 0 (5)  & 1 (0)    \\
dllama8b  & 0 (2) & 6 (0)    \\
llama70b & 0 (3)  & 0 (0)    \\
dllama70b  & 0 (2)  & 4 (0)  \\
qwen14b  & 0 (4)  & 6 (0)    \\
dqwen14b &  1 (1)  & 4 (1)   \\
qwen32b  & 0 (2)  & 3 (1)    \\
dqwen32b & 0 (2)  & 5 (0)    \\
\textbf{Total} & 1 (21)  & 29 (2)    \\\hline
white woman  & 0 (4)  & 6 (0)    \\
white man & 1 (5)  & 4 (0)    \\
asian woman &  0 (3)  & 4 (0)   \\
asian man & 0 (3)  & 6 (0)   \\
hispanic woman  & 0 (1)  & 0 (1)    \\
hispanic man &  0 (3)  & 5 (1)    \\
black woman &  0 (2)  & 2 (0)    \\
native man &  0 (0)  & 2 (0)   \\\hline
\end{tabular}
\end{center}

\caption{Out of 8 models or personas, how many times is the approach significantly better than best majority voting (how many times it is worse). This compares ensembling methods}
\label{tab:base_best_maj_voting} 
\end{table}

\begin{table}[t]
\begin{center}
\footnotesize
\begin{tabular}{l|r}\hline 
\multicolumn{1}{c|}{\textbf{Model/Persona}} & 
\multicolumn{1}{c}{\textbf{SVM}}  \\\hline

llama8b   & 5 (0)    \\
dllama8b & 7 (0)    \\
llama70b  & 2 (1)    \\
dllama70b    & 5 (0)  \\
qwen14b  &  8 (0)    \\
dqwen14b &   3 (1)   \\
qwen32b    & 4 (1)    \\
dqwen32b   & 7 (0)    \\ 
\textbf{Total}   & 41 (3)    \\\hline 
white woman    & 7 (1)    \\
white man   & 7 (0)    \\
asian woman   & 5 (0)   \\
asian man   & 7 (0)   \\
hispanic woman   & 3 (1)    \\
hispanic man &  5 (1)    \\
black woman &  3 (0)    \\
native man &   4 (0)   \\\hline
\end{tabular}
\end{center}

\caption{Out of 8 models or personas, how many times is the approach significantly better than theoretical optimal weighted majority voting (how many times it is worse). This compares ensembling methods}
\label{tab:base_theo_opt} 
\end{table}

\newpage

\section{F1 Score Comparison Tables}
\begin{table*}[t]
\begin{center}
\footnotesize
\begin{tabular}{l|r|r|r|r|r|r|r|r}\hline 
\multicolumn{1}{c|}{\textbf{Model-config}} & 
\multicolumn{1}{c|}{\textbf{W Wom.}} & 
\multicolumn{1}{c|}{\textbf{W Man}} & 
\multicolumn{1}{c|}{\textbf{A. Wom.}} & 
\multicolumn{1}{c|}{\textbf{A. Man}} & 
\multicolumn{1}{c|}{\textbf{H. Wom.}} & 
\multicolumn{1}{c|}{\textbf{H. Man}} & 
\multicolumn{1}{c|}{\textbf{B. Wom.}} & 
\multicolumn{1}{c}{\textbf{N. Man}} \\\hline
\multicolumn{9}{c}{\textbf{Reasoning Effect}} \\\hline
llama8b &0.63 &0.66 &0.58 &\textbf{0.65} &0.66 &\textbf{0.68} &0.62 &\textbf{0.74} \\
dllama8b &\textbf{0.65} &\textbf{0.68} &\textbf{0.59} &0.61 &\textbf{0.73 }&0.67 &\textbf{0.63} &0.66 \\\hline
llama70b &\textbf{0.75} &\textbf{0.76} &\textbf{0.68} &\textbf{0.74} & 0.77 &\textbf{0.80} &\textbf{0.70} &\textbf{0.82} \\
dllama70b &0.73 &0.75 &0.65 &0.68 &0.77 &0.76 &0.68 &0.77 \\\hline
qwen14b &0.70 &0.71 &0.62 &0.63 &\textbf{0.77} &0.71 &0.67 &0.75 \\
dqwen14b &\textbf{0.71} &\textbf{0.73} &\textbf{0.64} &\textbf{0.68 }&0.76 &\textbf{0.73} &\textbf{0.68} &\textbf{0.76} \\\hline
qwen32b &\textbf{0.71} &\textbf{0.74} &\textbf{0.63} &\textbf{0.66} &\textbf{0.77} &\textbf{0.74} &\textbf{0.67} &\textbf{0.75} \\
dqwen32b &0.70 &0.72 &0.62 &0.64 &0.76 &0.72 &0.66 &0.73 \\\hline
\multicolumn{9}{c}{\textbf{Persona Effect}} \\\hline
llama8b &0.63 &0.66 &0.58 &0.65 &0.66 &0.68 &0.62 &0.74 \\
llama8b-persona &\textbf{0.68} &\textbf{0.69} &0.63 &0.71 &\textbf{0.70} &\textbf{0.71} &\textbf{0.65} &\textbf{0.77} \\
llama8b-val\_prof &\textbf{0.69} &\textbf{0.69} &0.68 &0.75 &\textbf{0.67} &\textbf{0.73} &\textbf{0.65} &\textbf{0.81} \\\hline
dllama8b &\textbf{0.65} &0.68 &0.59 &0.61 &\textbf{0.73} &0.67 &0.63 &0.66 \\
dllama8b-persona &0.64 &0.68 &0.55 &0.58 &\textbf{0.73} &0.65 &0.60 &0.63 \\
dllama8b-val\_prof &\textbf{0.65} &\textbf{0.70} &\textbf{0.61} &\textbf{0.64} &0.71 &\textbf{0.68} &\textbf{0.67} &\textbf{0.68} \\\hline
llama70b &0.75 &0.76 &0.68 &0.74 &\textbf{0.77} &0.80 &0.70 &0.82 \\
llama70b-persona &\textbf{0.78} &\textbf{0.78} &0.73 &\textbf{0.82} &0.74 &\textbf{0.81} &\textbf{0.71} &0.84 \\
llama70b-val\_prof &0.77 &0.76 &\textbf{0.74} &0.81 &0.75 &\textbf{0.81} &\textbf{0.71} &\textbf{0.85} \\\hline
dllama70b &0.73 &0.75 &0.65 &0.68 &\textbf{0.77} &0.76 &0.68 &0.77 \\
dllama70b-persona &\textbf{0.75} &\textbf{0.77} &0.69 &0.78 &0.76 &0.76 &0.69 &0.79 \\
dllama70b-val\_prof &0.72 &\textbf{0.77} &\textbf{0.70 }&\textbf{0.79 }&0.76 &\textbf{0.81} &\textbf{0.70} &\textbf{0.87} \\\hline
qwen14b &0.70 &0.71 &0.62 &0.63 &\textbf{0.77} &0.71 &0.67 &0.75 \\
qwen14b-persona &\textbf{0.72} &0.74 &0.65 &0.70 &\textbf{0.77 }&\textbf{0.76} &\textbf{0.68} &0.75 \\
qwen14b-val\_prof &0.70 &\textbf{0.76} &\textbf{0.66} &\textbf{0.71 }&0.75 &0.73 &0.67 &\textbf{0.83} \\\hline
dqwen14b &\textbf{0.71} &0.73 &0.64 &0.68 &\textbf{0.76} &0.73 &0.68 &0.76 \\
dqwen14b-persona &\textbf{0.71} &\textbf{0.74} &0.63 &\textbf{0.70} &\textbf{0.76} &\textbf{0.74 }&0.66 &0.76 \\
dqwen14b-val\_prof &0.70 &\textbf{0.74} &\textbf{0.65} &0.69 &0.75 &\textbf{0.74} &\textbf{0.69} &\textbf{0.86 }\\\hline
qwen32b &0.71 &0.74 &0.63 &0.66 &\textbf{0.77} &0.74 &0.67 &0.75 \\
qwen32b-persona &\textbf{0.76} &0.77 &\textbf{0.69} &\textbf{0.77} &0.74 &\textbf{0.81 }&\textbf{0.70} &0.81 \\
qwen32b-val\_prof &0.74 &\textbf{0.78} &\textbf{0.69} &0.73 &0.75 &0.79 &\textbf{0.70} &\textbf{0.83} \\\hline
dqwen32b &0.70 &0.72 &0.62 &0.64 &\textbf{0.76} &0.72 &0.66 &0.73 \\
dqwen32b-persona &\textbf{0.71} &0.73 &0.65 &0.71 &0.75 &0.77 &0.67 &0.75 \\
dqwen32b-val\_prof &\textbf{0.71} &\textbf{0.76} &\textbf{0.66} &\textbf{0.72 }&0.75 &\textbf{0.78} &\textbf{0.68 }&\textbf{0.82 }\\\hline
 
 \hline

\multicolumn{9}{c}{\textbf{Automated Prompt Engineering}} \\\hline
lama8b-persona &0.68 &0.69 &0.63 &0.71 &0.70 &0.71 &0.65 &0.77 \\
llama8b-persona\_opt &\textbf{0.70} &\textbf{0.71} &\textbf{0.64} &\textbf{0.77} &\textbf{0.75} &\textbf{0.73} &\textbf{0.66} &\textbf{0.82} \\\hline
dllama8b-persona &\textbf{0.64} &0.68 &0.55 &0.58 &0.73 &0.65 &\textbf{0.60} &0.63 \\
dllama8b-persona\_opt &0.63 &\textbf{0.71} &\textbf{0.60} &\textbf{0.63} &0.73 &\textbf{0.67} &0.56 &\textbf{0.71} \\\hline
llama70b-persona &\textbf{0.78} &0.78 &\textbf{0.73} &0.82 &0.74 &0.81 &\textbf{0.71} &0.84 \\
llama70b-persona\_opt &0.77 &0.78 &0.65 &0.82 &\textbf{0.76} &\textbf{0.82} &0.70 &\textbf{0.86 }\\\hline
dllama70b-persona &0.75 &\textbf{0.77} &\textbf{0.69} &\textbf{0.78} &0.76 &\textbf{0.76} &\textbf{0.69} &0.79 \\
dllama70b-persona\_opt &\textbf{0.76} &0.75 &0.65 &0.76 &\textbf{0.77} &0.74 &0.68 &\textbf{0.87} \\\hline
qwen14b-persona &0.72 &0.74 &\textbf{0.65} &\textbf{0.70} &\textbf{0.77} &\textbf{0.76} &\textbf{0.68} &0.75 \\
qwen14b-persona\_opt &\textbf{0.73} &0.74 &0.64 &0.69 &0.76 &0.74 &0.67 &\textbf{0.79} \\\hline
dqwen14b-persona &0.71 &0.74 &0.63 &\textbf{0.70} &0.76 &\textbf{0.74} &0.66 &0.76 \\
dqwen14b-persona\_opt &\textbf{0.72} &0.74 &\textbf{0.69} &0.69 &\textbf{0.77} &0.71 &\textbf{0.68} &0.76 \\\hline
qwen32b-persona &\textbf{0.76} &0.77 &0.69 &0.77 &0.74 &\textbf{0.81} &\textbf{0.70} &0.81 \\
qwen32b-persona\_opt &0.71 &0.77 &0.69 &0.77 &\textbf{0.77} &0.80 &0.69 &\textbf{0.83} \\\hline
dqwen32b-persona &\textbf{0.71} &0.73 &0.65 &0.71 &0.75 &\textbf{0.77} &0.67 &0.75 \\
dqwen32b-persona\_opt &0.69 &\textbf{0.74} &0.65 &\textbf{0.72} &\textbf{0.76} &0.71 &\textbf{0.69} &\textbf{0.83} \\\hline

\end{tabular}
\end{center}
\caption{F1 scores of each configuration for each persona}
\label{tab:non_ensembling_results} 
\end{table*}

\begin{table*}[t]
\begin{center}
\footnotesize
\begin{tabular}{l|r|r|r|r|r|r|r|r}\hline 
\multicolumn{1}{c|}{\textbf{Model-config}} & 
\multicolumn{1}{c|}{\textbf{W Wom.}} & 
\multicolumn{1}{c|}{\textbf{W Man}} & 
\multicolumn{1}{c|}{\textbf{A. Wom.}} & 
\multicolumn{1}{c|}{\textbf{A. Man}} & 
\multicolumn{1}{c|}{\textbf{H. Wom.}} & 
\multicolumn{1}{c|}{\textbf{H. Man}} & 
\multicolumn{1}{c|}{\textbf{B. Wom.}} & 
\multicolumn{1}{c}{\textbf{N. Man}} \\\hline
llama8b &0.63 &0.66 &0.58 &0.65 &0.66 &0.68 &0.62 &0.74 \\
llama8b-persona &0.68 &0.69 &0.63 &0.71 &0.70 &0.71 &0.65 &0.77 \\
llama8b-val\_prof &0.69 &0.69 &\textbf{0.68} &0.75 &0.67 &0.73 &0.65 &0.81 \\
llama8b-persona\_opt &0.70 &0.71 &0.64 &\textbf{0.77} &\textbf{0.75} &0.73 &0.66 &0.82 \\
llama8b-best\_majority &0.70 &\textbf{0.72} &\textbf{0.68} &\textbf{0.77} &\textbf{0.75} &\textbf{0.74} &0.66 &\textbf{0.83} \\
llama8b-weighted\_maj &\textbf{0.71} &0.71 &0.66 &0.75 &0.72 &\textbf{0.74} &0.66 &0.82 \\
llama8b-weighted\_maj\_theo &\textbf{0.71} &0.71 &0.66 &0.75 &0.72 &\textbf{0.74} &0.66 &0.82 \\
llama8b-SVM &\textbf{0.71} &\textbf{0.72} &0.67 &0.76 &\textbf{0.75} &0.73 &\textbf{0.67} & 0.80 \\\hline

dllama8b &0.65 &0.68 &0.59 &0.61 &0.73 &0.67 &0.63 &0.66 \\
dllama8b-persona &0.64 &0.68 &0.55 &0.58 &0.73 &0.65 &0.60 &0.63 \\
dllama8b-val\_prof &0.65 &0.70 &0.61 &0.64 &0.71 &0.68 &\textbf{0.67} &0.68 \\
dllama8b-persona\_opt &0.63 &0.71 &0.60 &0.63 &0.73 &0.67 &0.56 &0.71 \\
dllama8b-best\_majority &0.65 &0.71 &0.61 &0.64 &\textbf{0.74} &0.68 &\textbf{0.67} &0.71 \\
dllama8b-weighted\_maj &0.65 &0.71 &0.59 &0.63 &\textbf{0.74} &0.68 &0.64 &0.69 \\
dllama8b-weighted\_maj\_theo &0.65 &0.71 &0.59 &0.63 &\textbf{0.74} &0.68 &0.64 &0.69 \\
dllama8b-SVM &\textbf{0.69} &\textbf{0.73} &\textbf{0.65} &\textbf{0.71} &\textbf{0.74} &\textbf{0.72} &0.66 &\textbf{0.76} \\\hline

llama70b &0.75 &0.76 &0.68 &0.74 &0.77 &0.80 &0.70 &0.82 \\
llama70b-persona &\textbf{0.78} &0.78 &0.73 &\textbf{0.82} &0.74 &0.81 &0.71 &0.84 \\
llama70b-val\_prof &0.77 &0.76 &\textbf{0.74} &0.81 &0.75 &0.81 &0.71 &0.85 \\
llama70b-persona\_opt &0.77 &0.78 &0.65 &\textbf{0.82} &0.76 &0.82 &0.70 &0.86 \\
llama70b-best\_majority &\textbf{0.78} &\textbf{0.79} &\textbf{0.74} &\textbf{0.82} &\textbf{0.78} &\textbf{0.83} &\textbf{0.72} &\textbf{0.87} \\
llama70b-weighted\_maj &\textbf{0.78} &0.78 &0.73 &\textbf{0.82 }&0.77 &0.82 &0.70 &0.85 \\
llama70b-weighted\_maj\_theo &\textbf{0.78} &0.78 &0.73 &\textbf{0.82} &0.77 &0.82 &0.70 &0.85 \\
llama70b-SVM &\textbf{0.78} &0.78 &\textbf{0.74} &\textbf{0.82} &\textbf{0.78} &0.82 &0.70 &0.85 \\\hline

dllama70b &0.73 &0.75 &0.65 &0.68 &\textbf{0.77} &0.76 &0.68 &0.77 \\
dllama70b-persona &0.75 &0.77 &0.69 &0.78 &0.76 &0.76 &0.69 &0.79 \\
dllama70b-val\_prof &0.72 &0.77 &0.70 &0.79 &0.76 &0.81 &\textbf{0.70} &0.87 \\
dllama70b-persona\_opt &0.76 &0.75 &0.65 &0.76 &\textbf{0.77} &0.74 &0.68 &0.87 \\
dllama70b-best\_majority &0.76 &0.77 &0.70 &0.79 &\textbf{0.77} &0.81 &\textbf{0.70 }&0.87 \\
dllama70b-weighted\_maj &0.75 &0.77 &0.69 &0.78 &\textbf{0.77} &0.78 &\textbf{0.70} &0.87 \\
dllama70b-weighted\_maj\_theo &0.75 &0.77 &0.69 &0.78 & \textbf{0.77} &0.78 &\textbf{0.70} &0.87 \\
dllama70b-SVM &\textbf{0.77} &\textbf{0.78} &\textbf{0.71} &\textbf{0.80} &\textbf{0.77} &\textbf{0.82} &\textbf{0.70} &\textbf{0.88} \\\hline

qwen14b &0.70 &0.71 &0.62 &0.63 &0.77 &0.71 &0.67 &0.75 \\
qwen14b-persona &0.72 &0.74 &0.65 &0.70 &0.77 &0.76 &0.68 &0.75 \\
qwen14b-val\_prof &0.70 &0.76 &0.66 &0.71 &0.75 &0.73 &0.67 &0.83 \\
qwen14b-persona\_opt &0.73 &0.74 &0.64 &0.69 &0.76 &0.74 &0.67 &0.79 \\
qwen14b-best\_majority &0.73 &0.76 &0.66 &0.71 &\textbf{0.78} &0.76 &0.68 &0.83 \\
qwen14b-weighted\_maj &0.72 &0.74 &0.65 &0.70 &\textbf{0.78} &0.75 &0.68 &0.80\\
qwen14b-weighted\_maj\_theo &0.72 &0.74 &0.65 &0.70 &\textbf{0.78} &0.75 &0.68 &0.80 \\
qwen14b-SVM &\textbf{0.75} &\textbf{0.77} &\textbf{0.69} &\textbf{0.75} &\textbf{0.78} &\textbf{0.79} &\textbf{0.69} &\textbf{0.85} \\\hline

dqwen14b &0.71 &0.73 &0.64 &0.68 &0.76 &0.73 &0.68 &0.76 \\
dqwen14b-persona &0.71 &0.74 &0.63 &0.70 &0.76 &0.74 &0.66 &0.76 \\
dqwen14b-val\_prof &0.70 &0.74 &0.65 &0.69 &0.75 &0.74 &0.69 &\textbf{0.86} \\
dqwen14b-persona\_opt &0.72 &0.74 &0.69 &0.69 &0.77 &0.71 &0.68 &0.76 \\
dqwen14b-best\_majority &0.72 &\textbf{0.76 }&\textbf{0.69} &0.73 &\textbf{0.78} &0.75 &0.69 &\textbf{0.86} \\
dqwen14b-weighted\_maj &0.72 &\textbf{0.76} &0.68 &0.70 &\textbf{0.78} &0.74 &0.69 &0.85 \\
dqwen14b-weighted\_maj\_theo &0.72 &\textbf{0.76} &0.68 &0.70 &\textbf{0.78} &0.74 &0.69 &0.85 \\
dqwen14b-SVM &\textbf{0.75} &\textbf{0.76} &0.68 &\textbf{0.76} &0.77 &\textbf{0.78}&\textbf{0.70} &\textbf{0.86 }\\\hline

qwen32b &0.71 &0.74 &0.63 &0.66 &\textbf{0.77} &0.74 &0.67 &0.75 \\
qwen32b-persona &\textbf{0.76} &0.77 &\textbf{0.69} &0.77 &0.74 &\textbf{0.81} &\textbf{0.70} &0.81 \\
qwen32b-val\_prof &0.74 &\textbf{0.78} &\textbf{0.69} &0.73 &0.75 &0.79 &\textbf{0.70} &0.83 \\
qwen32b-persona\_opt &0.71 &0.77 &\textbf{0.69} &0.77 &\textbf{0.77} &0.80 &0.69 &0.83 \\
qwen32b-best\_majority &\textbf{0.76} &\textbf{0.78 }&\textbf{0.69} &0.77 &\textbf{0.77} &\textbf{0.81} &\textbf{0.70} &0.83 \\
qwen32b-weighted\_maj &0.74 &0.77 &\textbf{0.69} &0.77 &\textbf{0.77} &\textbf{0.81 }&0.69 &0.82 \\
qwen32b-weighted\_maj\_theo &0.74 &0.77 &\textbf{0.69} &0.77 &\textbf{0.77} &\textbf{0.81} &0.69 &0.82 \\
qwen32b-SVM &\textbf{0.76} &\textbf{0.78} &\textbf{0.69} &\textbf{0.78} &\textbf{0.77} &0.79 &\textbf{0.70} &\textbf{0.86} \\\hline

dqwen32b &0.70 &0.72 &0.62 &0.64 &0.76 &0.72 &0.66 &0.73 \\
dqwen32b-persona &0.71 &0.73 &0.65 &0.71 &0.75 &0.77 &0.67 &0.75 \\
dqwen32b-val\_prof &0.71 &0.76 &0.66 &0.72 &0.75 &0.78 &0.68 &0.82 \\
dqwen32b-persona\_opt &0.69 &0.74 &0.65 &0.72 &0.76 &0.71 &\textbf{0.69} &0.83 \\
dqwen32b-best\_majority &0.71 &0.76 &0.66 &0.72 &\textbf{0.77} &0.78 &\textbf{0.69} &0.83 \\
dqwen32b-weighted\_maj &\textbf{0.73} &0.75 &0.66 &0.72 &\textbf{0.77} &0.76 &\textbf{0.69} &0.81 \\
dqwen32b-weighted\_maj\_theo &\textbf{0.73} &0.75 &0.66 &0.72 &\textbf{0.77} &0.76 &\textbf{0.69} &0.81 \\
dqwen32b-SVM &\textbf{0.73} &\textbf{0.77} &\textbf{0.68} &\textbf{0.78} &0.76 &\textbf{0.80} &\textbf{0.69} &\textbf{0.86} \\\hline

\end{tabular}
\end{center}
\caption{F1 scores of each configuration for each persona. This table compares all methods.}
\label{tab:all_results} 
\end{table*}

\end{document}